\documentclass[conference]{IEEEtran}
\IEEEoverridecommandlockouts
\usepackage{cite}
\usepackage{amsmath,amssymb,amsfonts}
\usepackage{algorithmic}
\usepackage{graphicx}
\usepackage{textcomp}
\usepackage{xcolor}
\def\BibTeX{{\rm B\kern-.05em{\sc i\kern-.025em b}\kern-.08em
    T\kern-.1667em\lower.7ex\hbox{E}\kern-.125emX}}
\begin{document}

\title{Olfactory Inertial Odometry: Sensor Calibration and Drift Compensation}

\author{\IEEEauthorblockN{Kordel K. France}
\IEEEauthorblockA{\textit{Dept. of Computer Science} \\
\textit{University of Texas at Dallas}\\
Richardson, TX, USA \\
kordel.france@utdallas.edu}
\and
\IEEEauthorblockN{Ovidiu Daescu}
\IEEEauthorblockA{\textit{Dept. of Computer Science} \\
\textit{University of Texas at Dallas}\\
Richardson, TX, USA \\
ovidiu.daescu@utdallas.edu}
\and
\IEEEauthorblockN{Anirban Paul}
\IEEEauthorblockA{\textit{Dept. of Biochemistry} \\
\textit{University of Texas at Dallas}\\
Richardson, TX, USA \\
anirban.paul@utdallas.edu}
\and
\IEEEauthorblockN{Shalini Prasad}
\IEEEauthorblockA{\textit{Dept. of Biochemistry} \\
\textit{University of Texas at Dallas}\\
Richardson, TX, USA \\
shalini.prasad@utdallas.edu}
}

\pagestyle{plain}

\maketitle

\begin{abstract}
Visual inertial odometry (VIO) is a process for fusing visual and kinematic data to understand a machine's state in a navigation task.
Olfactory inertial odometry (OIO) is an analog to VIO that fuses signals from gas sensors with inertial data to help a robot navigate by scent.
Gas dynamics and environmental factors introduce disturbances into olfactory navigation tasks that can make OIO difficult to facilitate.
With our work here, we define a process for calibrating a robot for OIO that generalizes to several olfaction sensor types.
Our focus is specifically on calibrating OIO for centimeter-level accuracy in localizing an odor source on a slow-moving robot platform to demonstrate use cases in robotic surgery and touchless security screening.
We demonstrate our process for OIO calibration on a real robotic arm and show how this calibration improves performance over a cold-start olfactory navigation task.
\end{abstract}


\section{Introduction}
\vspace{-1mm}

The most primitive forms of navigation in biological organisms arose from olfactory tracking of odour plumes and pheremone trails to a food source. 
Odour plumes are dynamic, change directions with wind shifts, and are highly subjected to environmental constraints such as temperature and humidity. 
The strength of the plume also slowly decays with time as air equalizes, making it difficult to identify the tail of the odour plume. 
Consequentially, successful scent-based navigation is dependent on appropriately quantifying uncertainty, and our work here focuses on demonstrating how the extrapolation of visual inertial odometry (VIO) techniques specifically helps with locating the plume source in machine olfaction tasks.


\begin{figure}
  \centering
  \includegraphics[width=85mm]{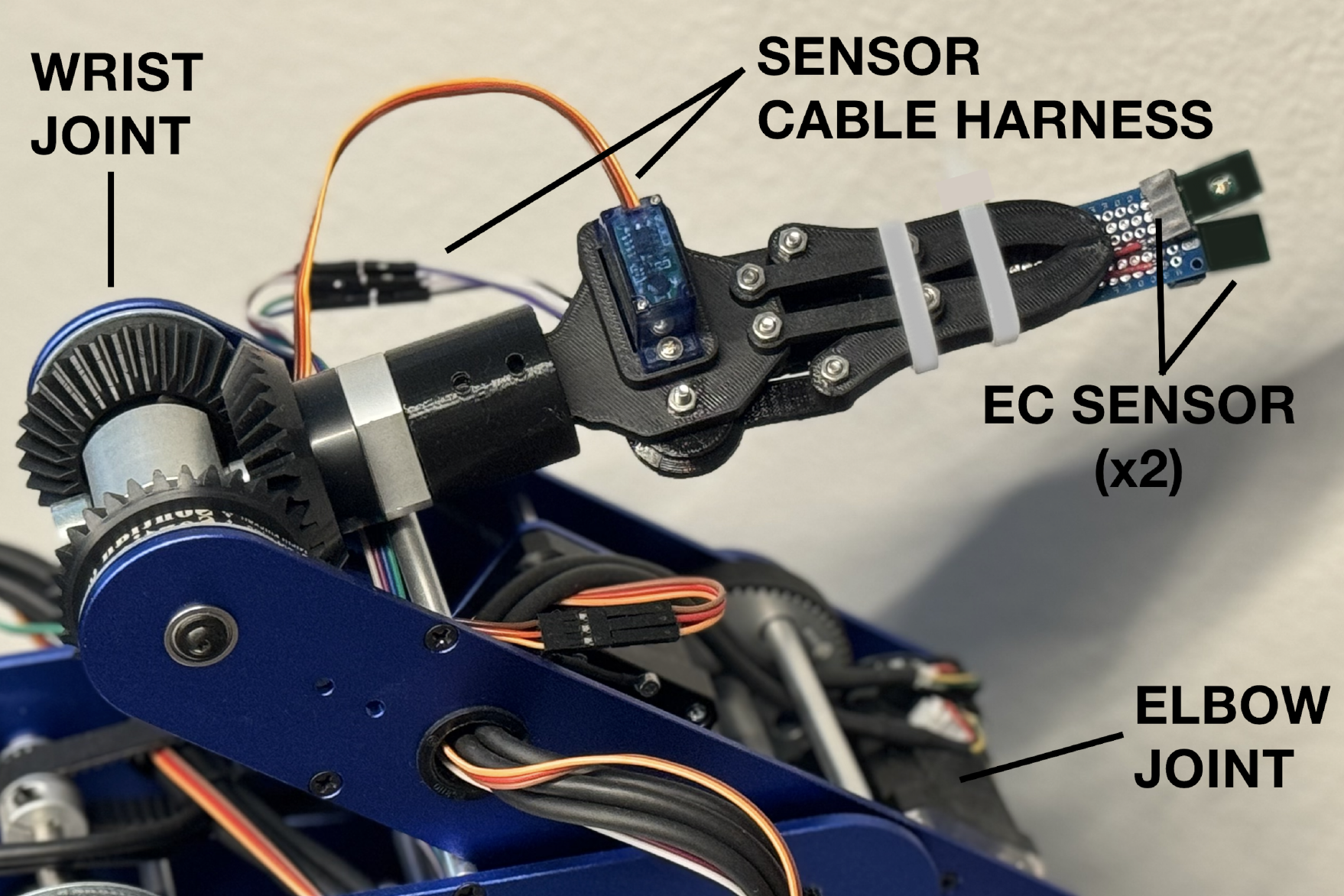}
  \caption{A diagram showing how the electrochemical sensor toolhead attaches to the robot "finger tip".}
  \label{fig:robot_ec_tool}
\end{figure}

\begin{figure}
  \centering
  \includegraphics[width=85mm]{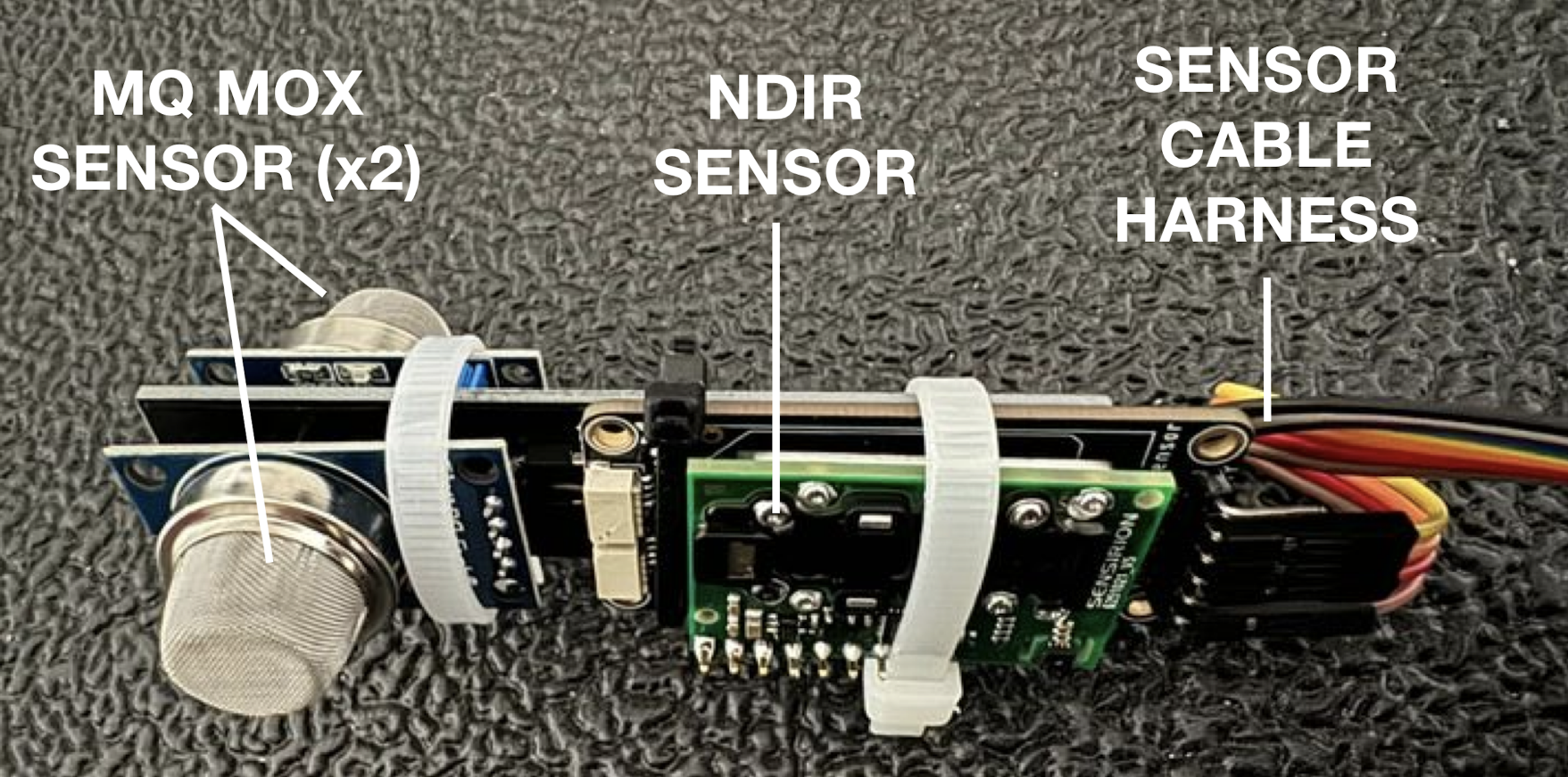}
  \caption{A diagram showing the MQ-series metal oxide sensor and NDIR sensor toolhead. An analogous toolhead is used for the MICS-series metal oxide sensors and photo-acoustic sensor.}
  \label{fig:robot_mox_tool}
\end{figure}

We define a protocol for calibrating a robot to facilitate scent-based navigation through the fusion olfaction and inertial measurements.
To show that our methodology is not specific to one sensor variety, we generalize our protocol over multiple olfaction sensors and various control algorithms.
The source of the plume must be tracked quickly because any scent, conditional on the air dynamics of the environment, will eventually equalize with the air, deleting the gradient that can be used for navigation.
Our goal is to show that careful sensor calibration and knowledge of noise priors can relieve the need for high-computation deep learning methods to obtain compelling results.
We present the theory behind our methods and then demonstrate it empirically with a real robotic arm tracking a scent via several olfaction sensors integrated onto the finger tip. 
Our intent is to model scenarios that can benefit robot-assisted surgery and touchless security screenings for explosives and drugs at airports.

In what follows, we denote our method as olfactory inertial odometry (OIO) and demonstrate that OIO can lead to desirable results in scent tracking.

Our contributions can be summarized as follows: 
\begin{enumerate}
\item We illustrate how visual inertial odometry techniques can be extrapolated to olfaction by fusing olfactory data with inertial data.
\item We demonstrate how OIO can be generalized to multiple olfaction sensor types.
\item We define a process for calibrating OIO to understand future drift and hysterisis distributions.
\end{enumerate}

\section{Background}
\vspace{-1mm}
One common pain point among machine olfaction research is the difference in sampling times needed for an olfaction sensor to detect (or react) with the target compound. 
These sampling times can take on the order of seconds, minutes, or even hours, and can change depending on the target compound.
This is a stark contrast from visual techniques with cameras that can measure at hundreds of frames per second.

Work from Dennler, et al. in \cite{dennler24} show how metal oxide sensors can react on the order of milliseconds through custom circuitry and bout detection methods from \cite{schmuker16-bout-detection}.
However, they do not demonstrate their methods over robotic navigation tasks and do not use inertial data.
A survey from \cite{burgues20-drone-chem-sense-survey} gives a great review on the state-of-the-art in olfactory navigation, but all mentioned applications use different navigation techniques all leverage a quadcopter.

Research from Burgues, et al. in \cite{sniffybug-single-burgues19} show how small quadcopters can localize the source of an odour, a project that was later extended by \cite{sniffybug-swarm-duisterhof21} illustrating the concept within swarms.
\cite{France2024} show considerations for swarming in groups of olfaction robots, but their methodology is only simulated and only addresses electrochemical sensors.
With all three works, they leverage a single type of olfaction sensor and do not allude to VIO techniques for navigation.
We extend their work to other sensor varieties and attempt to demonstrate how OIO can enable centimeter-level odour localization on a slower robotic platform (a robotic arm) versus meter-level localization on a faster robotic platform (a quadcopter).
This sacrifice in platform speed for detection accuracy through the use of a robot arm more accurately mirrors our intended applications for robot-assisted surgery and touchless security screenings.
Hassan, et al. in \cite{hassan_llm_multimodal_olfaction_2024} construct a scent-guided robot by fusing visual and olfactory data with a large language model. 
However, this method does not run at the edge and is dependent on high-computation deep learning methods.

Other research, such as the work of Singh et al. in \cite{Singh2023} and \cite{ScentienceApp}, show how plume tracking via scent can be made tractable, but they demonstrate only in simulation and rely heavily on deep learning. 
A difference in our work is to demonstrate scent-based navigation on a real robot, and how it can be done effectively without a heavy reliance on deep learning.
We define a foundational calibration procedure as a VIO analog to olfactory navigation on which deep learning can later be built.

\section{Methodology}
\vspace{-1mm}

To show that OIO generalizes over different olfaction sensors and sample times, we demonstrate a calibration technique over five different sensor types: electrochemical (EC) sensors, two variants of metal oxide (MOX) sensors, photoacoustic (PA) sensors, and non-dispersive infrared (NDIR) sensors.

Each set of sensors are integrated into the "finger tip" of the robotic arm. 
These sensors "sniff" the air in search of the target compound.
Previous work from \cite{dennler22drift} and \cite{PaulPrasad2020} find that MOX and EC sensors, respectively, can give different drift and heteroscedasticity for identical sensor types due to the nature of their underlying technologies.
To counter this, we use two EC sensors and two MOX sensors for each MOX sensor type.
One sensor is enabled, while the other is disabled, and this oscillatory process repeats the entire time that the robot scans for a signal. 
The signals between both of these sensors are simply averaged here, but a future work will investigate how the gradient between both sensors can inform the direction in which the robot will move.
A diagram of the EC sensor toolhead attached to the robot is shown in Figure \ref{fig:robot_ec_tool}, while the toolhead for the NDIR and MQ-series MOX sensors is shown in Figure \ref{fig:robot_mox_tool}.
We place them on a 100mm extending board to maximize passive airflow over the sensor.


With our robotic arm, we have access to inertial kinematics in each joint which allow us to fuse inertial odometry techniques with the olfactory signal, analogous to how vision and inertial data are fused for VIO.
Fusion of the olfactory signal and inertial data occurs through the use of extended Kalman filters (EKF).
Encoder drift values per limb are all less than $1e^{-3}$ deg / sec$^2$ according to the robot manufacturer.

In order to calibrate OIO, we perform two processes:
\begin{enumerate}
    \item Calibrate the measurement error for each degree of freedom (DoF) according to sensor type in Section B. This establishes our Type A uncertainty.
    \item Calibrate the measurement error for all degrees of freedom combined through a wireless network inspired algorithm in Section C. This establishes our Type B uncertainty.
\end{enumerate}
In Section D, we discuss how these uncertainty values derived from calibration are used to inform the process noise covariance $Q$ and sensor noise covariance $R$ for our EKFs. 

\subsection{Sensor Bout Detection}
We start by calibrating one degree of freedom.
The calibration occurs \textit{after} the warm-up period has finished for each sensor type.
We then sample five initial measurements, $[y_0, ... , y_4]$.
We leverage the bout detection methods from Schmuker, et al. in \cite{schmuker16-bout-detection}.
They establish their method for metal oxide sensors, but we leverage the same principles for all five sensor types with some refinements suggested by \cite{dennler22drift}.

To filter the olfactory signal and provide smoother gradient following, we define a window $k$ of length $5$ around each of the raw measurements $y_t$ and compute the moving average $y'_t$ at each time step.
Each measurement $y'$ at time $t-1$ is subtracted from the measurement $y_t$ at the current time $t$, effectively measuring the differential of $y_t$: 

\begin{equation}
    \delta_t = y'_t - y'_{t-1}
    \label{eq:boutDelta}
\end{equation}

If the differential $\delta_t$ exceeds the previous value $\delta_{t-1}$ \textit{and} the current absolute value $y'_t$ is also greater than the initial baseline measurements taken at the beginning of the experiment $[y_0, ... , y_4]$, we assume the robot is moving toward the source of the odour. 
We find that, for optical-based sensors such as the NDIR and PA sensors, we can leverage a lower value for $k$.
However, to standardize the overall process, we generalize a $k$ for all sensors.
Optimal windows for each DoF are 3, 5, 7, and 11 for 1-, 2-, 3-, and 5-DoF, respectively.

The above process is repeated for 2-, 3-, and 5-DoF.
A summary of the error for each sensor for each DoF is defined in Table \ref{tab:timeTable}.
The values for these sensor errors are then used to define the EKF for the whole robot calibration process detailed in the next section.

\subsection{Calibrating Type A Uncertainty}
We begin by tracking a target scent via one DoF.
The robot is only allowed to move the shoulder in azimuth.
This limits the robot's ability to only use a casting technique to locate the plume.
With this 1-DoF, we effectively construct a 3-armed bandit problem where the only possible actions are move left, move right, or do nothing.
Then we successively increase to  2-DoF (azimuth and elevation in the shoulder), 3-DoF (azimuth and elevation in the shoulder plus tilt in the elbow), and then 5-DoF (azimuth and elevation in the shoulder, tilt at the elbow, and both tilt and roll at the wrist.).
With  2-DoF, we effectively enable two-dimensional search of the target compound.
Expansion to 3-DoF enables the robot to surge toward the target compound, effectively enabling 3D search.
5-DoF grants us the ability to perform angle-level measurements in a 3D search space.
By building up DoF in this manner, we follow the work of Eschmann, et al. in \cite{eschmann2023learning} where they demonstrate the importance of building the simulation complexity toward increasing degrees of freedom, versus starting with the target degrees all at once.
We calculate the standard deviation $s$ over our bout detection measurement window $k$, and use these values to construct our Type A uncertainty value $u_a$ for each DoF according to the following:

\begin{equation}
    u_a = \frac{s}{\sqrt{k}}
    \label{eq:uncertaintyTypeA}
\end{equation}

\subsection{Calibrating Type B Uncertainty via Belief Maps}

\begin{table*}
\caption{Total Time for Calibration}
\begin{center}
\begin{tabular}{|c|c|c|c|c|c|}
\hline
\textbf{Sensor}& \textbf{Manufacturer}& $\tau_{lower}$ (s)& $\tau_{upper}$ (s)& \textbf{Total Time (s)}& \textbf{Error (PPM)}\\
\hline
NDIR& Sensirion& 0.1& 1.0& 51& 30 \\
PA& Sensirion& 0.2& 1.0& 55& 50\\
EC& Texas Instruments& 0.5& 6.0& 87& 20\\
MOX (MQ)& Olimex& 0.5& 3.0& 89& 100\\
MOX (MICS)& SGX Sensortech& 0.1& 3.0& 71& 100\\
\hline
\end{tabular}
\label{tab:timeTable}
\end{center}
\end{table*}

\begin{figure}
  \centering
  \includegraphics[width=75mm]{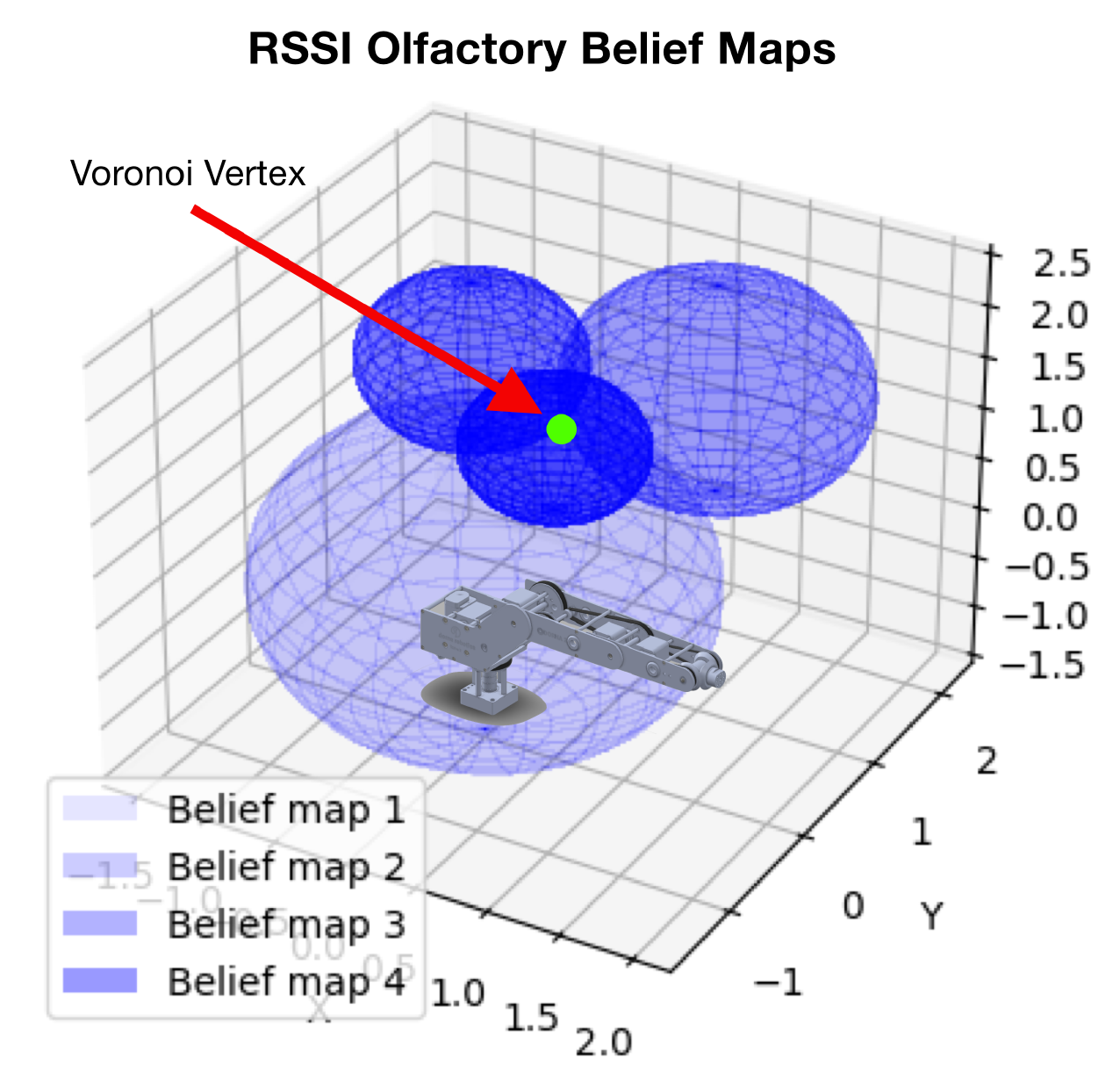}
  \caption{Four belief map spheres intersecting to form the Voronoi vertex.}
  \label{fig:rssi_belief_maps}
\end{figure}

In the field of signal processing, the received signal strength indicator (RSSI) serves as a widely used metric for path loss, helping to determine the proximity of a receiver to a transmitter. 
We adapt a technique from signal processing to convert the responses from olfactory sensors into RSSI values. 

Figure \ref{fig:rssi_belief_maps} denotes a visual for the following process.
We utilize a signal processing technique to transform olfactory signals into RSSI belief maps, indicating potential plume locations based on the robot arm's movements. Each RSSI value represents the radius of a sphere, with points on its surface, known as "sigma points," having equal probabilities of being the target compound's location. By maneuvering the sensor, we can resample from a new position, creating a second sphere that intersects with the first, resulting in a circle that confines potential locations to its circumference. As the robot continues to move and establish additional spheres, the intersections of these spheres significantly reduce the probability space for the plume source to just two sigma points. Ultimately, if all spheres intersect at a single point, they do so at the Voronoi vertex, representing the odour's location.

The preceding process creates a belief map of potential plume source locations through trilateration, and repeates until the Voronoi vertex is found. 
In a perfect situation, only 4 movements with 5 measurements would be necessary to pinpoint the source of the target compound. 
However, in practice, several environmental influences complicate this process, as plumes are highly dynamic and can change over time, leading to considerable backtracking and requiring more movements. 

We take the Euclidean distance from the Voronoi vertex to the true location of the odour source +/-$v$ and the total number of moves needed to locate the Voronoi vertex $m$ and use this to inform our Type B uncertainty as follows:

\begin{equation}
    u_b = \frac{2v}{\sqrt{m}}
    \label{eq:uncertaintyTypeB}
\end{equation}

\begin{figure*}
  \centering
  \includegraphics[width=\textwidth]{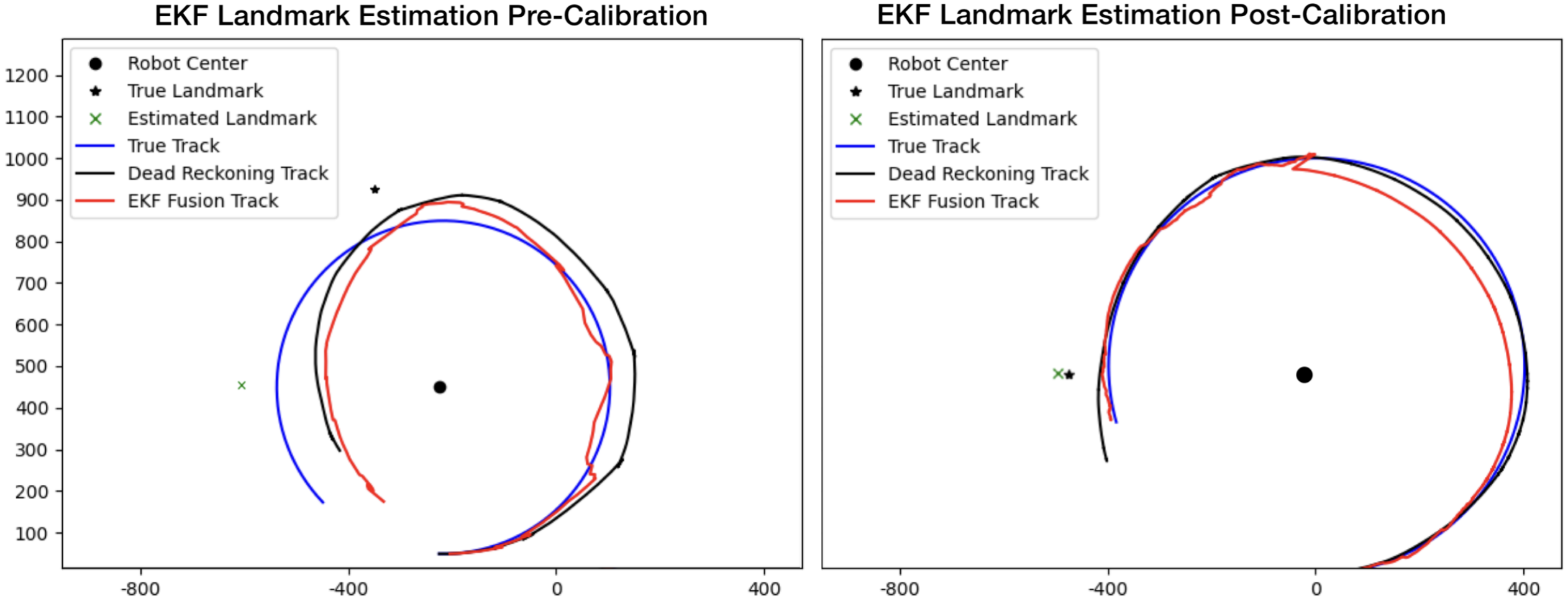}
  \caption{Results for EKF Pre- and Post-Calibration. Axis units are in millimeters.}
  \label{fig:expEkf}
\end{figure*}

\section{Results}
\vspace{-1mm}
Every EKF has a process noise covariance $Q$ and a sensor noise covariance noise $R$.
We place our uncertainties $u_a$ and $u_b$ along the diagonals of $R$ and $Q$ respectively. 
These are then used to tune our EKF for fusion of the sensor signal and robot inertial data.
In some instances, particularly for EC and MOX sensors, we find that some additional fine-tuning of the parameters may be valuable, but value is marginal.

A diagram of the final EKF produced from the OIO calibration process for the EC sensors is shown in Figure \ref{fig:expEkf}, while a summary of the final calibration times for all sensors are shown in Table \ref{tab:timeTable}.
$\tau_{lower}$ and $\tau_{upper}$ define the acceptable lower and upper sampling times of each sensor, respectively.

\section{Conclusion}
\vspace{-1mm}
The above process delineates a process that we have found effective for calibrating for OIO.
We have attempted to define a strategy that can generalize to multiple olfaction sensor types, but there is still much to be proven.
Future work will define additional experimental controls, address more sensor types, and demonstrate the use of reinforcement learning and deep learning autoencoders to enable further value in sensor fusion.

In consideration of our results, we find evidence to support that effective and accurate navigation via scent can be accomplished through the fusion of inertial and olfaction data.
In addition, we enumerate careful considerations in the simulation design and construction of the reward signal to elicit better translation of results from simulation to reality.
We hope this inspires more research in machine olfaction and expands awareness about the potential for olfactory navigation.

\vspace{-1mm}
\centering
\bibliographystyle{IEEEtran}
\bibliography{sample-base}

\begin{thebibliography}{10}
\providecommand{\url}[1]{#1}
\csname url@samestyle\endcsname
\providecommand{\newblock}{\relax}
\providecommand{\bibinfo}[2]{#2}
\providecommand{\BIBentrySTDinterwordspacing}{\spaceskip=0pt\relax}
\providecommand{\BIBentryALTinterwordstretchfactor}{4}
\providecommand{\BIBentryALTinterwordspacing}{\spaceskip=\fontdimen2\font plus
\BIBentryALTinterwordstretchfactor\fontdimen3\font minus \fontdimen4\font\relax}
\providecommand{\BIBforeignlanguage}[2]{{%
\expandafter\ifx\csname l@#1\endcsname\relax
\typeout{** WARNING: IEEEtran.bst: No hyphenation pattern has been}%
\typeout{** loaded for the language `#1'. Using the pattern for}%
\typeout{** the default language instead.}%
\else
\language=\csname l@#1\endcsname
\fi
#2}}
\providecommand{\BIBdecl}{\relax}
\BIBdecl

\bibitem{dennler24}
\BIBentryALTinterwordspacing
N.~Dennler, D.~Drix, T.~P.~A. Warner, S.~Rastogi, C.~D. Casa, T.~Ackels, A.~T. Schaefer, A.~van Schaik, and M.~Schmuker, ``High-speed odor sensing using miniaturized electronic nose,'' \emph{Science Advances}, vol.~10, no.~45, p. eadp1764, 2024. [Online]. Available: \url{https://www.science.org/doi/abs/10.1126/sciadv.adp1764}
\BIBentrySTDinterwordspacing

\bibitem{schmuker16-bout-detection}
\BIBentryALTinterwordspacing
M.~Schmuker, V.~Bahr, and R.~Huerta, ``Exploiting plume structure to decode gas source distance using metal-oxide gas sensors,'' \emph{Sensors and Actuators B: Chemical}, vol. 235, pp. 636--646, 2016. [Online]. Available: \url{https://www.sciencedirect.com/science/article/pii/S0925400516307833}
\BIBentrySTDinterwordspacing

\bibitem{burgues20-drone-chem-sense-survey}
\BIBentryALTinterwordspacing
J.~Burgués and S.~Marco, ``Environmental chemical sensing using small drones: A review,'' \emph{Science of The Total Environment}, vol. 748, p. 141172, 2020. [Online]. Available: \url{https://www.sciencedirect.com/science/article/pii/S004896972034701X}
\BIBentrySTDinterwordspacing

\bibitem{sniffybug-single-burgues19}
\BIBentryALTinterwordspacing
J.~Burgués, V.~Hernández, A.~J. Lilienthal, and S.~Marco, ``Smelling nano aerial vehicle for gas source localization and mapping,'' \emph{Sensors}, vol.~19, no.~3, 2019. [Online]. Available: \url{https://www.mdpi.com/1424-8220/19/3/478}
\BIBentrySTDinterwordspacing

\bibitem{sniffybug-swarm-duisterhof21}
B.~P. Duisterhof, S.~Li, J.~Burgués, V.~J. Reddi, and G.~C. H.~E. de~Croon, ``Sniffy bug: A fully autonomous swarm of gas-seeking nano quadcopters in cluttered environments,'' in \emph{2021 IEEE/RSJ International Conference on Intelligent Robots and Systems (IROS)}, 2021, pp. 9099--9106.

\bibitem{France2024}
\BIBentryALTinterwordspacing
K.~K. France, A.~Paul, I.~Banga, and S.~Prasad, ``Emergent behavior in evolutionary swarms for machine olfaction,'' in \emph{Proceedings of the Genetic and Evolutionary Computation Conference}, ser. GECCO '24.\hskip 1em plus 0.5em minus 0.4em\relax New York, NY, USA: Association for Computing Machinery, 2024, p. 30–38. [Online]. Available: \url{https://doi.org/10.1145/3583131.3590376}
\BIBentrySTDinterwordspacing

\bibitem{hassan_llm_multimodal_olfaction_2024}
\BIBentryALTinterwordspacing
S.~Hassan, L.~Wang, and K.~R. Mahmud, ``Integrating vision and olfaction via multi-modal llm for robotic odor source localization,'' \emph{Sensors}, vol.~24, no.~24, 2024. [Online]. Available: \url{https://www.mdpi.com/1424-8220/24/24/7875}
\BIBentrySTDinterwordspacing

\bibitem{Singh2023}
\BIBentryALTinterwordspacing
S.~H. Singh, F.~van Breugel, R.~P.~N. Rao, and B.~W. Brunton, ``Emergent behaviour and neural dynamics in artificial agents tracking odour plumes,'' \emph{Nature Machine Intelligence}, vol.~5, no.~1, pp. 58--70, Jan 2023. [Online]. Available: \url{https://doi.org/10.1038/s42256-022-00599-w}
\BIBentrySTDinterwordspacing

\bibitem{ScentienceApp}
\BIBentryALTinterwordspacing
K.~K. France, ``Scentience,'' \url{https://apps.apple.com/us/app/scentience/id6741092923}, 2025, accessed: 2025-03-02. [Online]. Available: \url{https://apps.apple.com/us/app/scentience/id6741092923}
\BIBentrySTDinterwordspacing

\bibitem{dennler22drift}
\BIBentryALTinterwordspacing
N.~Dennler, S.~Rastogi, J.~Fonollosa, A.~{van Schaik}, and M.~Schmuker, ``Drift in a popular metal oxide sensor dataset reveals limitations for gas classification benchmarks,'' \emph{Sensors and Actuators B: Chemical}, vol. 361, p. 131668, 2022. [Online]. Available: \url{https://www.sciencedirect.com/science/article/pii/S0925400522003100}
\BIBentrySTDinterwordspacing

\bibitem{PaulPrasad2020}
\BIBentryALTinterwordspacing
A.~Paul, S.~Muthukumar, and S.~Prasad, ``Review—room-temperature ionic liquids for electrochemical application with special focus on gas sensors,'' \emph{Journal of The Electrochemical Society}, vol. 167, no.~3, p. 037511, dec 2019. [Online]. Available: \url{https://dx.doi.org/10.1149/2.0112003JES}
\BIBentrySTDinterwordspacing

\bibitem{eschmann2023learning}
J.~Eschmann, D.~Albani, and G.~Loianno, ``Learning to fly in seconds,'' 2023.

\end{thebibliography}

\end{document}